\documentclass{article}
\usepackage{spconf,amsmath,graphicx}
\usepackage{svg}
\usepackage{tipa}
\usepackage{colortbl}

\title{Layer-Wise Analysis of Self-Supervised Acoustic Word Embeddings: \\ A Study on Speech Emotion Recognition}
\name{Alexandra Saliba$^*$\thanks{$^*$Equal contribution.}, Yuanchao Li$^*$, Ramon Sanabria, Catherine Lai}
\address{University of Edinburgh}
\begin{document}
%
\maketitle
\begin{abstract}
The efficacy of self-supervised speech models has been validated, yet the optimal utilization of their representations remains challenging across diverse tasks. In this study, we delve into Acoustic Word Embeddings (AWEs), a fixed-length feature derived from continuous representations, to explore their advantages in specific tasks. AWEs have previously shown utility in capturing acoustic discriminability. In light of this, we propose measuring layer-wise similarity between AWEs and word embeddings, aiming to further investigate the inherent context within AWEs. Moreover, we evaluate the contribution of AWEs, in comparison to other types of speech features, in the context of Speech Emotion Recognition (SER). Through a comparative experiment and a layer-wise accuracy analysis on two distinct corpora, IEMOCAP and ESD, we explore differences between AWEs and raw self-supervised representations, as well as the proper utilization of AWEs alone and in combination with word embeddings. Our findings underscore the acoustic context conveyed by AWEs and showcase the highly competitive SER accuracies by appropriately employing AWEs.

\end{abstract}
\begin{keywords}
HuBERT, acoustic word embeddings, self-supervised speech models, speech emotion recognition\end{keywords}

\section{INTRODUCTION}
\label{sec:intro}

Unlike discrete embeddings of word tokens, speech representations are presented in a continuous form with variable lengths, which can impede their utility in certain scenarios such as query-by-example, voice search, keyword spotting, and word discovery \cite{zhang2009unsupervised,barakat2011keyword,de2007template}. Therefore, fixed-dimensional speech representations, specifically Acoustic Word Embeddings (AWEs), have been employed to capture acoustic information in variable-length speech segments \cite{maas2012word}. This process condenses the information in a manner that guarantees segments containing the same words are mapped to similar embeddings. Previous studies have demonstrated that the use of AWEs allows for the application of simple calculations, such as cosine similarity, to measure the distance between different speech segments, as opposed to using computationally expensive methods like dynamic time warping~\cite{levin_2013}.

Moreover, owing to the contextualized nature of self-supervised representations, a simple mean-pooling approach is sufficient to construct AWEs that capture acoustic discriminability which implicitly contains word meanings and sequential information \cite{sanabria_analyzing_2023}. To this end, we aim to investigate \textit{whether the self-supervised AWEs outperform their continuous form (i.e., the raw frame-level representations)} and \textit{whether they present similar contextual meanings to their corresponding word embeddings}. To achieve this goal, we use Speech Emotion Recognition (SER) as the downstream task for the following reasons:

\textbf{1)} The variable length of frame-level speech representations has posed a longstanding challenge for SER. To standardize the input length, SER usually suffers from either noise introduced by padding or information loss caused by chopping \cite{lin2021chunk}. Thanks to the fixed length, we hypothesize that \textit{AWEs serve as an ideal solution to address this problem, especially when use in combination with word embeddings.}

\textbf{2)} SER usually requires both acoustic and lexical information for satisfactory performance. However, the inter-modal incongruity sometimes makes it challenging to combine the two input sources \cite{wang2023crossattention}. As AWEs can discriminate words compared to continuous speech representations, we hypothesize that \textit{AWEs align better with word embeddings via proper fusion approaches (e.g., cross-attention) due to their discretized characteristics.}

To validate our hypotheses, in this work, we first compute the layer-wise similarity between AWEs and word embeddings. Subsequently, we build an SER system utilizing AWEs from a self-supervised model HuBERT as input and compare its performance with the raw HuBERT representations, as well as with Mel spectrogram baselines. Additionally, we incorporate BERT embeddings through simple concatenation and cross-attention for comparison, aiming to further understand how AWEs differ from raw self-supervised representations.

\section{Related Work}
\label{sec:relatedwork}

\subsection{Self-Supervised AWEs}
Self-supervised speech models have been intensively studied in recent years and have proven effective in various speech tasks \cite{yang2021superb}. Building upon this foundation, there has been an emerging exploration of self-supervised AWEs. Sanabria et al. \cite{sanabria_analyzing_2023} compared AWEs constructed from wav2vec 2.0 \cite{baevski_wav2vec_2020}, XLSR-53 \cite{conneau2020unsupervised}, and HuBERT \cite{hsu_hubert_2021}. They noted that HuBERT AWEs by mean-pooling rival the SOTA on English AWEs, and despite being trained only on English, HuBERT AWEs evaluated on Xitsonga, Mandarin, and French consistently outperform the multilingual model XLSR-53 as well as wav2vec 2.0 trained on English.

\subsection{SER via Self-Supervised Representations}
\label{sec:layerwise_analysis}

Prior studies have explored the utilization of self-supervised representations in SER. Li et al. \cite{li_exploration_2023} conducted a thorough investigation of wav2vec 2.0 across different layers and observed that representations extracted from the middle layer lead to the best SER performance. Furthermore, they confirmed the absence of paralinguistic information in wav2vec 2.0 representations. However, with two simple linear layers, these representations are still sufficient to achieve satisfactory results. Additionally, Morais et al. \cite{morais2022speech} utilized wav2vec 2.0 and HuBERT for SER and demonstrated the effectiveness of both of the two models. 

Despite the aforementioned progress of AWEs and SER with self-supervised learning, a question persists: \textit{How do self-supervised AWEs differ from the raw self-supervised representations?} With this question in mind, we follow the path of previous work, aiming to investigate self-supervised AWEs to provide insights into the optimal utilization of pre-trained self-supervised models in downstream tasks.

\section{Corpora and Self-Supervised AWEs}

\subsection{Corpora Description}

We adopt two distinct corpora: IEMOCAP \cite{busso_iemocap_2008} and ESD \cite{zhou_emotional_2022}. IEMOCAP comprises five dyadic sessions involving ten actors (five male and five female), each engaging in scripted and improvised multimodal interactions. The corpus contains approximately 12 hours of speech annotated by three annotators with ten emotion classes. Consistent with prior research \cite{li_exploration_2023}, we merged \textit{Happy} and \textit{Excited}, and excluded utterances without transcripts. This process results in a total of 5,500 utterances utilized in this study, each assigned one label from the following four classes: \textit{Angry, Happy, Neutral, and Sad}.

ESD consists of 17,500 utterances spoken by 10 native English actors, each labeled with one of five different emotions (\textit{Angry, Surprise, Sad, Happy, Neutral}). A notable feature is the inclusion of parallel data: \textit{pairs of utterances with the same content and speaker but different emotions.} This design enables control for linguistic content, facilitating the isolation of emotion-related acoustic information, thereby resulting in higher SER accuracies.

\subsection{HuBERT AWEs}
We use \textit{HuBERT-base-ls960}\footnote{https://huggingface.co/facebook/hubert-base-ls960}, which has a 7-layer convolutional encoder, followed by a BERT-like encoder with 12 Transformer layers, and a projection layer. The CNN encoder uses 25ms windows with a 20ms frame rate. The masking is applied to the CNN's output, and the masked sequence is passed on to the BERT-like encoder. 

We downsample acoustic data to 16kHz and pass it through HuBERT to extract frame-level representations from each of the 12 layers (denoted as L1-L12), as well as from the output of the CNN (denoted as L0). Subsequently, we obtain word time steps using the Montreal Forced Aligner \cite{mcauliffe_montreal_2017}. Following \cite{sanabria_analyzing_2023}, we apply mean-pooling over the frames of each word to obtain the AWEs as the mean-pooling has been proven highly effective compared to sophisticated learned pooling approaches \cite{peng_correspondence_2020}.

\section{Experiments}
\subsection{Similarity Between AWEs and Lexical Embeddings}
\label{sec:4.1}

In previous research on the layer-wise analysis of self-supervised speech models, Pasad et al. \cite{pasad_layer-wise_2021, pasad_comparative_2023} demonstrated that these models encode speech information, progressing from the frame-level to phonetic-level representations and then to word identity and meaning. Li et al. \cite{li_exploration_2023}, Lin et al. \cite{lin_utility_2022}, and Zhu et al. \cite{zhu2023deep} applied the representations to prosody-related tasks, confirming the distinctive contributions of different layers. In this study, however, we compare HuBERT AWEs with BERT\footnote{https://huggingface.co/bert-base-uncased} embeddings, aiming to identify the layers of AWEs that are most sensitive to linguistic information and to explore the contextual distribution of AWEs.

To this end, we compute the Local Neighborhood Similarity (LNS) \cite{boggust_embedding_2022} of a word $w$ between the lexical and acoustic embedding spaces. The underlying idea behind this practice is that \textit{if the feature vectors of a given word in two different embedding spaces encode similar contextual information, they will share common local neighbors.} The LNS of a word $w$ is determined by the similarity between the set of $K$ nearest neighbors in the  HuBERT AWEs ($HBA$) and BERT word embeddings ($BERT$), respectively:
\begin{align}
LNS(w) = J(KNN_{HBA}(w), KNN_{BERT}(w))
\end{align}
where Jaccard distance (intersection over union) between two sets is defined as:
\begin{align}
J(C_{1}, C_{2}) = \frac{C_{1}\cap C_{2}}{C_{1}\cup C_{2}}
\end{align}

\begin{figure}
    \centering
    \includegraphics[width=\columnwidth]{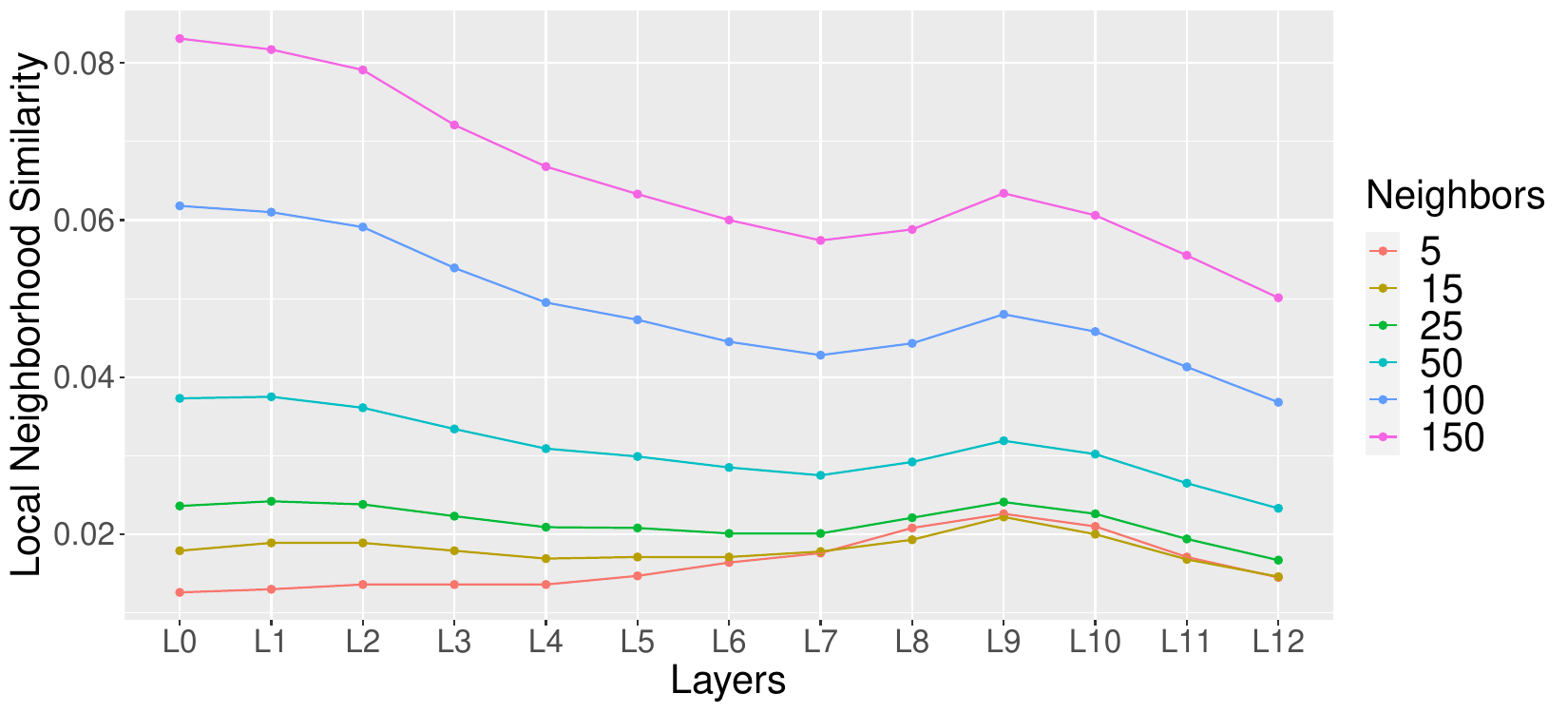}
    \caption{LNS per layer with different numbers of neighbors.}
    \label{fig:jaccard_IEMOCAP}
\vspace{-5pt}
\end{figure}

As an identical pattern is obtained on ESD, we present the results only on IEMOCAP for brevity, which are plotted in Fig.~\ref{fig:jaccard_IEMOCAP}. It is evident that the similarity between acoustic and lexical word embeddings is consistently low, typically ranging between 1\% and 2.5\%. Across all layers, there is a noticeable peak at layer 9, suggesting that layer 9 encodes acoustic word meanings that more align with the information encoded by the language model, which is consistent with \cite{pasad_comparative_2023}. Even though, it is worth noting that the value remains low.

Moreover, we extract the 5 closest acoustic and lexical neighbors for the words \textit{anyhow} and \textit{after} from layer 9 as examples to illustrate the dissimilarity between the two distributions. As seen in Table~\ref{tab:neighbor}, while there is one word in common between the two sets of neighbors, acoustic neighbors often share a common vowel sound with the target word, while lexical neighbors tend to be more syntactically related. Alongside the findings from Fig.~\ref{fig:jaccard_IEMOCAP}, it is noted that, despite AWEs being able to acoustically discriminate words \cite{sanabria_analyzing_2023}, they demonstrate a distribution distinct from that of word embeddings in terms of context, which we refer to as \textit{acoustic context}.

\begin{table}[ht]
\centering
\caption{Example of 5 closest acoustic and lexical neighbors for the words \textit{anyhow} and \textit{after} from layer 9.}
\begin{tabular}{l|l|l} \hline 
\textbf{Word} & \textbf{Lexical Neighbors} & \textbf{Acoustic Neighbors} \\ \hline 
anyhow & \textbf{anyway}, didn't, say, & \textbf{anyway}, while, now,\\
 & it, well & annie, why \\ \hline 
after & be, a, \textbf{at}, & \textbf{at}, chapter, asked, \\ 
 & give, i & act, grasshopper \\ \hline
\end{tabular}
\label{tab:neighbor}
\vspace{-5pt}
\end{table}

\subsection{SER Experiments}
\subsubsection{Experimental Settings}

To delve deeper into the distinct characteristics of AWEs, we conduct a downstream task—SER, using AWEs as input. Given that SER leverages contributions from both acoustic and lexical information, and following a prior study that investigated the use of raw self-supervised representations \cite{li_exploration_2023}, we perform a similar approach to explore the differences between AWEs and raw representations from HuBERT.

We train a simple neural network classifier with two dense hidden layers (sizes 128 and 16), employing ReLU activation, as sophisticated networks are deemed unnecessary when utilizing powerful self-supervised representations for downstream tasks \cite{li_exploration_2023, yang2021superb}. As our objective is to self-contrast for novel findings rather than compare with the literature, we conduct an 80/20 train/test split instead of cross-validation. Results are measured using Weighted Accuracy (WA).

\subsubsection{Comparative Study}
We compare three types of acoustic features: Mel spectrograms, raw HuBERT representations, and HuBERT AWEs (referred to as Mel, HuBERT, and AWEs for brevity). For the latter two, we compute the accuracy from every layer and report the average score. Additionally, we explore their fusion with BERT embeddings through concatenation and cross-attention (audio and BERT features attend to each other and are then combined) using the best-performing layer. Note that the feature dimension of Mel differs from that of BERT. Although it is possible to encode Mel to the same dimension as that of BERT for cross-attention, the performance is significantly inferior to HuBERT and AWEs. Thus, for brevity, we only report concatenation for Mel. Each model is trained five times, and we report the average score.

The results are presented in Table~\ref{tab:IEMOCAP_accuracies} and \ref{tab:ESD_accuracies}, and the following findings are observed. \textbf{1)} Concatenating BERT embeddings significantly improves the performance of Mel. However, the contribution of such lexical information is less pronounced when integrated into HuBERT or AWEs. This observation highlights a previous finding that self-supervised speech features already convey implicit lexical information compared to raw speech features \cite{sanabria_analyzing_2023,pasad_layer-wise_2021,li_exploration_2023}, rendering BERT embeddings somewhat redundant.

\textbf{2)} On IEMOCAP: The performance using AWEs is slightly inferior to that of using HuBERT, which is reasonable due to information loss caused by mean-pooling during AWEs construction. While AWEs contain context, which may contribute as additional information, such an advantage does not manifest. On the contrary, although AWEs still exhibit lower performance than HuBERT when fused with BERT through cross-attention, the gap between them diminishes (only 0.04). This phenomenon can be attributed to discretized acoustic features (i.e., AWEs) being more easily aligned with word embeddings compared to continuous ones through cross-attention, a mechanism designed to capture relatedness between two representations. This alignment naturally leads to higher enhancement.

\begin{table}[ht]
\vspace{-5pt}
\centering
\caption{SER accuracy on IEMOCAP.}
\begin{tabular}{lc}
\hline
\textbf{Model Input} & \textbf{WA (Std) \%} \\ \hline
Mel & 54.89 ($\pm$0.68) \\
Mel+BERT (concat.) & 66.19 ($\pm$1.29) \\
HuBERT & 72.43 ($\pm$0.12) \\
HuBERT+BERT (concat.) & 73.98 ($\pm$0.18) \\
HuBERT+BERT (cross-att.) & \textbf{74.44} ($\pm$0.31) \\
AWEs & 72.12 ($\pm$0.53) \\
AWEs+BERT (concat.) & 73.61 ($\pm$0.36) \\
AWEs+BERT (cross-att.) & 74.40 ($\pm$0.44) \\
\hline
\end{tabular}
\label{tab:IEMOCAP_accuracies}
\end{table}

\begin{table}[ht]
\centering
\caption{SER accuracy on ESD.}
\begin{tabular}{lc}
\hline
\textbf{Model Input} & \textbf{WA (Std) \%} \\ \hline
Mel & 70.11 ($\pm$0.59) \\
Mel+BERT (concat.) & 62.11 ($\pm$0.57) \\
HuBERT & 91.81 ($\pm$0.08) \\
HuBERT+BERT (concat.) & 90.98 ($\pm$0.11) \\
HuBERT+BERT (cross-att.) & 92.50 ($\pm$0.09) \\
AWEs & \textbf{93.20} ($\pm$0.13) \\
AWEs+BERT (concat.) & 88.80 ($\pm$0.19) \\
AWEs+BERT (cross-att.) & 92.21 ($\pm$0.13) \\
\hline
\end{tabular}
\label{tab:ESD_accuracies}
\end{table}

\textbf{3)} On ESD: Concatenating BERT embeddings decreases performance compared to using acoustic features alone. On the other hand, cross-attention enhances performance for HuBERT, yet has the opposite effect for AWEs. These phenomena are attributed to the fact that sentences in ESD remain unchanged for every emotion, having no contribution to emotion discrimination. Concatenating BERT embeddings only brings redundancy, thereby decreasing performance. Nonetheless, cross-attention with BERT enhances performance as it models the relatedness between the source and target features. Unlike concatenation, which uses the two directly as input, the same BERT embeddings contribute differently to SER based on different acoustic features serving as the query in cross-attention. Moreover, we observe that when using acoustic features alone, AWEs outperform HuBERT, contrary to the findings on IEMOCAP. Given that sentences in ESD remain consistent for every emotion, this setup facilitates the acoustic context (explained in Sec.~\ref{sec:4.1}) as an indicator for SER. In contrast, sentences in IEMOCAP vary, making the acoustic context less predictable and challenging to leverage.

\subsubsection{Layer-Wise Accuracy Analysis}
\begin{figure}[ht]
    \centering
    \includegraphics[width=\columnwidth]{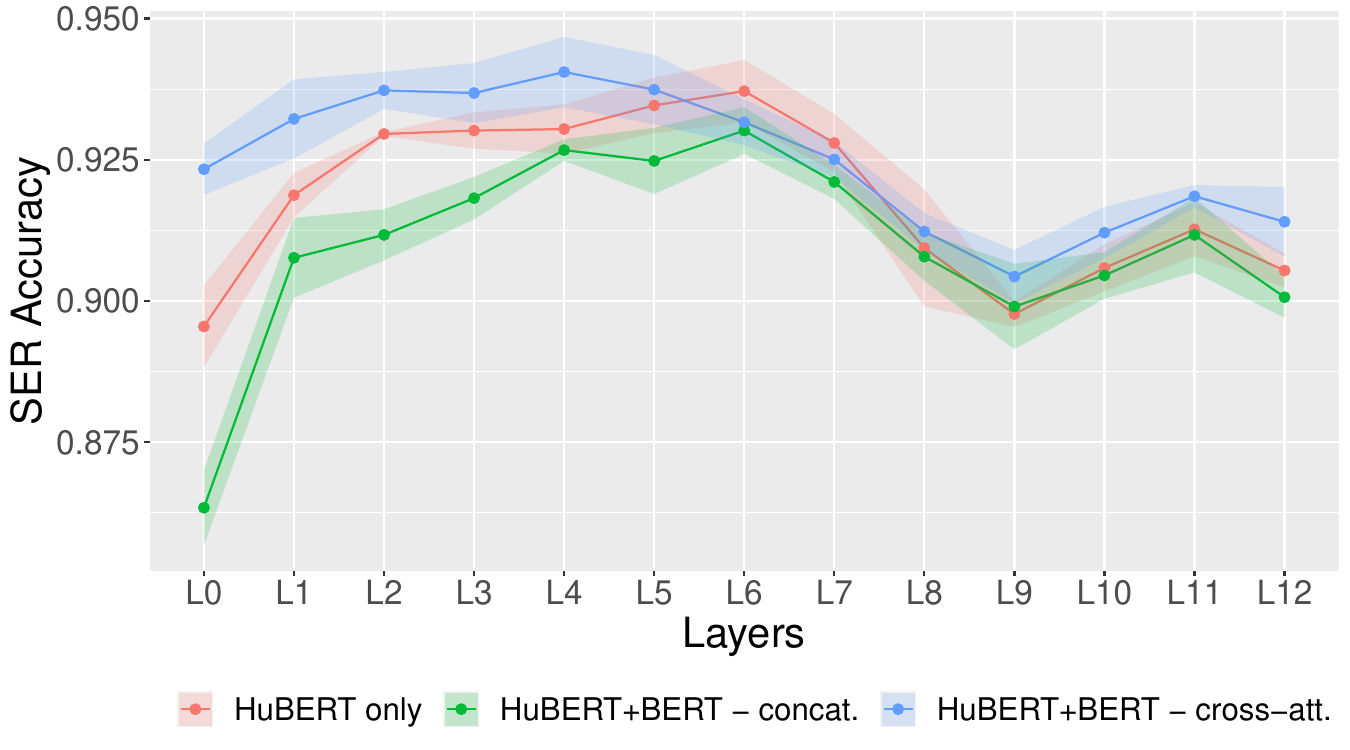}
     \caption{Layer-wise analysis of SER accuracy of HuBERT w/ and w/o fusion of BERT embeddings on ESD.}
    \label{fig:ESD_plot_comparison_HuBERT}
\end{figure}

\begin{figure}[ht]
    \centering
    \includegraphics[width=\columnwidth]{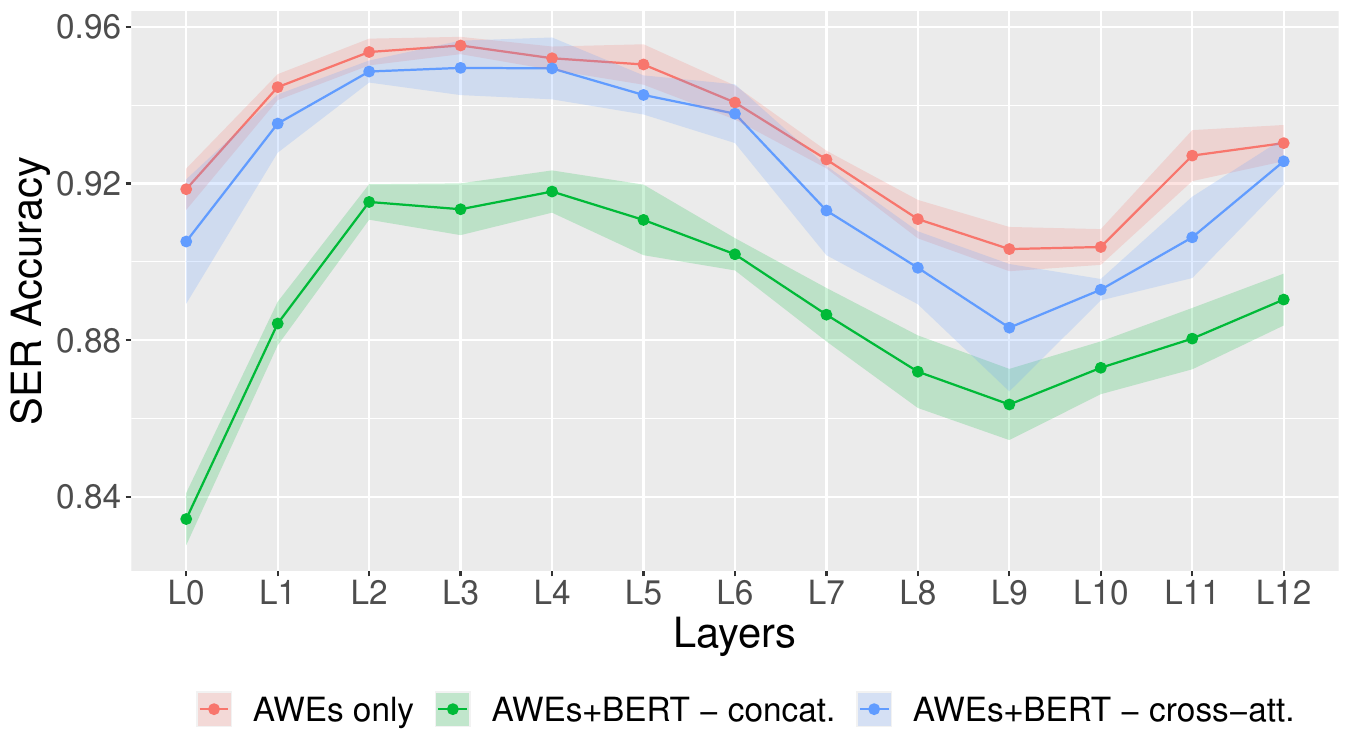}
     \caption{Layer-wise analysis of SER accuracy of AWEs w/ and w/o fusion of BERT embeddings on ESD.}
    \label{fig:ESD_plot_comparison_AWEs}
\end{figure}

Subsequently, we perform a layer-wise accuracy analysis, illustrating mean and std scores in Fig.~\ref{fig:ESD_plot_comparison_HuBERT} and \ref{fig:ESD_plot_comparison_AWEs}. As similar layer-wise patterns have been found in both corpora, we omit IEMOCAP for brevity (the major differences have been described in the last section with Table~\ref{tab:IEMOCAP_accuracies}).

By comparing the two figures, we can observe several differences between HuBERT raw representations and AWEs: \textbf{1)} Using AWEs alone consistently outperforms the two fusions, yet using HuBERT alone underperforms cross-attention at most layers. \textbf{2)} On the shallow layers, AWEs have significant advantages over raw HuBERT, and the optimal layer for using AWEs is around layer 3 instead of the middle layer. \textbf{3)} When using raw HuBERT, the discrepancy among the three input types diminishes after the middle layer. This is because the deeper layers start encoding lexical information, which holds little value in ESD. In contrast, when using AWEs, the discrepancy persists, showcasing the advantage of the acoustic context represented by AWEs in situations where lexical information has minimal impact. \textbf{4)} When using AWEs, the last layer does not exhibit a drop that commonly observed in raw self-supervised representations \cite{pasad_layer-wise_2021,li_exploration_2023,pasad_comparative_2023}. Instead, the performances rise, offering a solution for more effectively leveraging the last layer representations.

\section{Conclusion}
In this paper, we conduct a layer-wise analysis of AWEs derived from HuBERT. Specifically, we measure the similarity between AWEs and BERT embeddings, discovering that AWEs exhibit a distinct context distribution from word embeddings, oriented toward the acoustic aspect. Furthermore, through a comparative study on two distinct corpora, we demonstrate the advantages of AWEs over other types of speech features. Alongside a layer-wise accuracy analysis, we unveil the relationship between AWEs and raw HuBERT representations, as well as the practice of utilizing AWEs alone and in combination with word embeddings. Our findings are expected to provide insights into the realm of self-supervised speech models and inspire future research on better utilizing self-supervised representations in different speech tasks.

\small
\bibliographystyle{IEEEbib}
\bibliography{refs}

\end{document}